\documentclass[letterpaper]{article} 
\makeatletter
\def\copyright@text{}
\makeatother

\usepackage{aaai2026}  
\renewcommand{\footnoterule}{\kern -3pt}
\usepackage{times}  
\usepackage{helvet}  
\usepackage{courier}  
\usepackage[hyphens]{url}  
\usepackage{graphicx} 
\def\UrlFont{\rm}  
\usepackage{natbib}  
\usepackage{caption} 
\usepackage{array}
\usepackage{algorithm}
\usepackage{algorithmic}
\usepackage{comment}
\usepackage{amsmath}

\usepackage{newfloat}
\usepackage{listings}
\DeclareCaptionStyle{ruled}{labelfont=normalfont,labelsep=colon,strut=off} 
\floatstyle{ruled}
\newfloat{listing}{tb}{lst}{}
\floatname{listing}{Listing}
\title{LLM4Sweat: A Trustworthy Large Language Model for Hyperhidrosis Support}
\author{
    Wenjie Lin,
    Jin Wei-Kocsis
}
\affiliations{
    
    School of Applied and Creative Computing\\
    Purdue University\\

    \{lin1790, kocsis0\}@purdue.edu
}

\usepackage{bibentry}
\usepackage{booktabs} 
\usepackage{multirow}

\begin{document}

\maketitle

\begin{abstract}
While large language models (LLMs) have shown promise in healthcare, their application for rare medical conditions is still hindered by scarce and unreliable datasets for fine-tuning. Hyperhidrosis, a disorder causing excessive sweating beyond physiological needs, is one such rare disorder, affecting 2-3\% of the population and significantly impacting both physical comfort and psychosocial well-being. To date, no work has tailored LLMs to advance the diagnosis or care of hyperhidrosis. To address this gap, we present LLM4Sweat, an open-source and domain-specific LLM framework for trustworthy and empathetic hyperhidrosis support. The system follows a three-stage pipeline. In the data augmentation stage, a frontier LLM generates medically plausible synthetic vignettes from curated open-source data to create a diverse and balanced question-answer dataset. In the fine-tuning stage, an open-source foundation model is fine-tuned on the dataset to provide diagnosis, personalized treatment recommendations, and empathetic psychological support. In the inference and expert evaluation stage, clinical and psychological specialists assess accuracy, appropriateness, and empathy, with validated responses iteratively enriching the dataset. Experiments show that LLM4Sweat outperforms baselines and delivers the first open-source LLM framework for hyperhidrosis, offering a generalizable approach for other rare diseases with similar data and trustworthiness challenges.
\end{abstract}


\section{Introduction}
\label{Introduction}

\begin{figure*}[ht]
  \includegraphics[width=\linewidth]{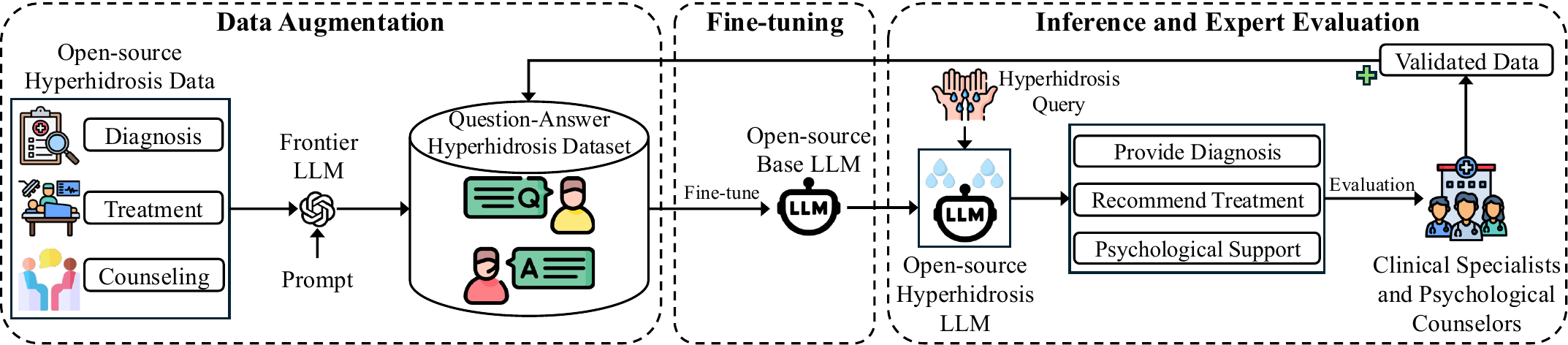}
  \caption {Overview of our proposed LLM4Sweat framework.}
  \label{fig:overview}
\end{figure*}

Hyperhidrosis, characterized by excessive sweating beyond thermoregulatory needs, affects approximately 2-3\% of the population and can significantly impair both physical comfort and psychosocial well-being~\cite{strutton2004epidemiology, solish2007impact}. Current therapeutic protocols span from potent topical antiperspirants to advanced procedural interventions such as botulinum toxin injections, microwave thermolysis (miraDry), iontophoresis, and endoscopic thoracic sympathectomy~\cite{hong2012clinical, gregoriou2019management,martinez2024endoscopic}. Despite the availability of these treatments, reliable and personalized medical support for diagnosis, treatment selection, and psychosocial management remains limited and unregulated~\cite{parashar2023impact}. Notably, the majority of cases represent primary hyperhidrosis, which is inherited and characterized by focal symptoms and has been linked to several genetic loci~\cite{henning2019genetic}.

Beyond clinical progress, parallel efforts in engineering have sought to improve diagnosis and daily management. Examples include polymer‐based insole designs using additive manufacturing for plantar hyperhidrosis management \cite{camargo2024polymer}, rectangular patch antenna electrolyte sensors for sweat composition analysis \cite{sakhawat2024patch}, and a wearable system for continuous sweat level monitoring \cite{bellapukonda2023logistic}. Also, daily management devices are proposed, alongside systematic engineering reviews of diagnostic and intervention technologies~\cite{lin2022systematic}, illustrating the potential for integrating engineering innovations into hyperhidrosis care. However, they remain largely decoupled from intelligent reasoning systems that can deliver holistic diagnosis, treatment guidance, and psychosocial support.

LLMs have shown promising capabilities in a range of healthcare applications, including diagnosis support \cite{singhal2023large}, treatment planning \cite{zhao2025autonomous}, and patient counseling \cite{qiu2024interactive}. However, their application adapted to hyperhidrosis and similar rare disorders remains largely absent. Unlike common conditions with abundant datasets, hyperhidrosis suffers from scarce high-quality data and fragmented, unreliable online information. These gaps hinder the trustworthy use of LLMs in clinical care and underscore the need for data-centric frameworks that can turn limited resources into reliable medical knowledge.

To address these limitations, we present LLM4Sweat, the first open-source LLM framework designed for trustworthy hyperhidrosis support. It overcomes the aforementioned challenges through a three-stage pipeline. Our work makes three key contributions: (1) we construct the hyperhidrosis question-answer dataset from open-source data and generate medically plausible synthetic vignettes by a frontier LLM; (2) we introduce a closed-loop fine-tuning and expert-in-the-loop evaluation pipeline that adapts open-source lightweight LLM to perform three integrated tasks, provide diagnosis, recommend treatment, and psychological support, within a unified framework; and (3) experiments show that our framework outperforms baselines and demonstrates strong potential for clinical deployment. Beyond addressing a critical gap in LLM adaptation for hyperhidrosis, it also offers a transferable solution to tackle data scarcity and trustworthiness challenges across other rare medical conditions.

\section{Related Work}
\label{Related Work}

\subsection{Clinical Diagnosis and Treatment of Hyperhidrosis}

The diagnosis of hyperhidrosis is primarily clinical, relying on patient history and physical examination, though these assessments are often subjective in practice~\cite{henning2021diagnose}. Once identified, treatment typically follows a stepwise approach, beginning with topical aluminum chloride hexahydrate and advancing to oral anticholinergics, iontophoresis, botulinum toxin injections, or device-based options such as miraDry~\cite{hong2012clinical, gregoriou2019management}. For severe and refractory cases, surgical intervention with endoscopic thoracic sympathectomy may be considered, although it carries a significant risk such as compensatory sweating~\cite{martinez2024endoscopic}.

\subsection{Engineering Solutions for Hyperhidrosis}
Recent work has highlighted engineering-driven diagnostic and monitoring innovations. Sakhawat et al.\ designed an antenna-based electrolyte sensor using Whatman filter paper to detect sodium and chloride concentrations in sweat for palmoplantar hyperhidrosis patients \cite{sakhawat2024patch}. 3D-printed thermoplastic polyurethane insoles have been developed to manage plantar hyperhidrosis by improving comfort and reducing moisture retention \cite{camargo2024polymer}. Hyun et al.\ leveraged machine learning models to classify hyperhidrosis types and predict compensatory hyperhidrosis level after surgery with high accuracy \cite{hyun2023machine}.  Bellapukonda et al.\ applied logistic regression to predict sweat levels with high accuracy based on environmental, physiological, and demographic variables, \cite{bellapukonda2023logistic}. Lin and Fang \cite{lin2022systematic} provided a comprehensive review of current and potential hyperhidrosis interventions from both medical and engineering perspectives, proposing smart wearable devices for daily management.

\subsection{Synthetic Data in Healthcare}
Synthetic data generation has emerged as a solution for medical domains with limited annotated datasets~\cite{peng2023study}. GAN-based models have also been used to expand rare disease datasets~\cite{kazeminia2020gans}. Diffusion models have been successfully applied to dermatological imaging tasks, producing high-fidelity synthetic skin lesion data to improve classifier robustness~\cite{farooq2024derm}. For text-based healthcare tasks, LLM-driven augmentation has improved performance in low-resource clinical NLP applications~\cite{barr2025large}. 

\subsection{LLMs in Medical Applications}
Recent work has shown that state-of-the-art LLMs such as GPT and Llama achieve competitive performance in clinical reasoning, treatment recommendation, and patient communication \cite{singhal2023large}. LLMs have also been used for empathetic patient support \cite{alanezi2024assessing}. Psychological distress is also a key challenge in rare conditions such as hyperhidrosis~\cite{solish2007impact}, underscoring the need for both accurate medical guidance and empathetic support. 

Building on these advances, our framework leverages a frontier LLM to generate medically plausible vignettes from open-source hyperhidrosis-related data, converting scarce and noisy real-world resources into a structured and balanced dataset. This enriched dataset enables reliable fine-tuning, equipping the model to provide trustworthy diagnostic reasoning, personalized treatment recommendations, and empathetic counseling. Furthermore, in the inference and expert evaluation stage, outputs by the fine-tuned model from users' queries are reviewed by clinical and psychological specialists, creating a feedback loop that further enhances accuracy, appropriateness, and trustworthiness.

\section{Methodology}
\label{Methodology}
Figure~\ref{fig:overview} provides an overview of the LLM4Sweat framework, which comprises closed-loop three stages: data augmentation, fine-tuning, and inference with expert-in-the-loop evaluation.

\begin{table*}[ht]
\centering
\begin{tabular}{m{2.3cm}|cccc|cccc|cccc}
\toprule
\center \multirow{2}{*}{Model} & \multicolumn{4}{c|}{Diagnosis} & \multicolumn{4}{c|}{Treatment} & \multicolumn{4}{c}{Overall} \\
\cmidrule(lr){2-5} \cmidrule(lr){6-9} \cmidrule(lr){10-13}
 & Acc & Prec & Rec & F1 & Acc & Prec & Rec & F1 & Acc & Prec & Rec & F1 \\
\midrule
 Llama-3.2-1B  & 0.425 & 0.595 & 0.434 & 0.425 & 0.425 & 0.340 & 0.380 & 0.340 & 0.425 & 0.543 & 0.421 & 0.403 \\
\midrule
Llama-3.2-1B (\textbf{LLM4Sweat}) \textbf{\small w/o Expert Eval} & \textbf{0.900} & \textbf{0.944} & \textbf{0.865} & \textbf{0.887} & \textbf{0.725} & \textbf{0.736} & \textbf{0.770} & \textbf{0.735} & \textbf{0.813} & \textbf{0.830} & \textbf{0.816} & \textbf{0.807} \\
\midrule
Llama-3.2-1B (\textbf{LLM4Sweat}) & \textbf{0.925} & \textbf{0.942} & \textbf{0.909} & \textbf{0.918} & \textbf{0.825} & \textbf{0.848} & \textbf{0.842} & \textbf{0.830} & \textbf{0.875} & \textbf{0.899} & \textbf{0.880} & \textbf{0.879} \\
\midrule
 Llama-3.2-3B  & 0.475 & 0.798 & 0.491 & 0.569 & 0.700 & 0.917 & 0.683 & 0.753 & 0.588 & 0.856 & 0.587 & 0.670 \\
\midrule
Llama-3.2-3B (\textbf{LLM4Sweat}) \textbf{\small w/o Expert Eval} & \textbf{0.825} & \textbf{0.821} & \textbf{0.828} & \textbf{0.819} & \textbf{0.825} & \textbf{0.902} & \textbf{0.822} & \textbf{0.856} & \textbf{0.825} & \textbf{0.861} & \textbf{0.827} & \textbf{0.839} \\
\midrule
 Llama-3.2-3B (\textbf{LLM4Sweat}) & \textbf{0.875} & \textbf{0.883} & \textbf{0.850} & \textbf{0.861} & \textbf{0.925} & \textbf{0.941} & \textbf{0.917} & \textbf{0.923} & \textbf{0.900} & \textbf{0.902} & \textbf{0.898} & \textbf{0.900} \\
\bottomrule
\end{tabular}
\caption{Comparison of our LLM4Sweat framework against baseline models for hyperhidrosis diagnosis and treatment task performance. We report accuracy (Acc), precision (Prec), recall (Rec), and F1 scores (F1) across models.}
\label{results}
\end{table*}

\subsection{Data Augmentation Stage}
We begin with a self-curated set of open-source hyperhidrosis-related data
\begin{equation}
  \mathcal{D}_{\text{real}} = \{(q_i, a_i, T_i^j)\}_{i=1}^N  
\end{equation}
where $q_i$ is a patient query, $a_i$ is the corresponding answer, and $T_i^j$ denotes the task label with $j \in \{1,2,3\}$ corresponding to three tasks \emph{diagnosis}, \emph{treatment}, and \emph{counseling}, respectively. $\mathcal{D}_{\text{real}}$ is for testing the fine-tuned models.

To address the low-resource nature of hyperhidrosis-specific data, we use a frontier LLM $f_{\theta_0}$ with carefully designed prompts to generate a vigenette synthetic dataset:
\begin{equation}
    \mathcal{D}_{\text{syn}} = \{(q'_k, a'_k, T_k^j)\}_{k=1}^M \sim f_{\theta_0}
\end{equation}
where $q'_k$ denotes synthetic queries in vignettes and $a'_k$ represents the synthetic answers. They form synthetic question–answer pairs for one of the defined task categories $T^j_k$. Besides, we ensure balanced generation so that the three task types $T^1$, $T^2$, and $T^3$ are proportionally represented in $\mathcal{D}_{\text{syn}}$.

\subsection{Fine-Tuning Stage}
We fine-tune an open-source foundation model $g_{\theta}$ to the hyperhidrosis domain via supervised fine-tuning on $\mathcal{D}_{\text{syn}}$. The training objective maximizes the conditional likelihood of the correct answer given the question and task:
\[
\theta^* = \arg\max_{\theta} \sum_{(q,a,T^j) \in \mathcal{D}_{\text{syn}}} \log p_\theta(a \mid q, T^j)
\]

\subsection{Inference and Expert Evaluation Stage}
At inference time, a user query $q^{\text{user}}$ is assigned a task label $T^{j,\text{pred}}$ based on the query context. The fine-tuned model $g_{\theta^*}$ generates a response:
\[
\hat{a} = g_{\theta^*}(q^{\text{user}}, T^{j,\text{pred}}),
\]

All generated outputs are validated by clinical specialists and psychological counselors for accuracy, appropriateness, and empathy. Validated output $\hat{a}_v$ will be fed into the synthetic dataset $\mathcal{D}_{\text{syn}}$ created in the data augmentation stage to further fine-tune $g_{\theta^*}$. This stage creates a dynamic training loop where the fine-tuned model is iteratively refined with validated expert feedback, improving both factual correctness and trustworthiness over successive cycles.

\section{Experiments}
\label{Experiments}

\begin{figure*}[ht]
  \includegraphics[width=\linewidth]{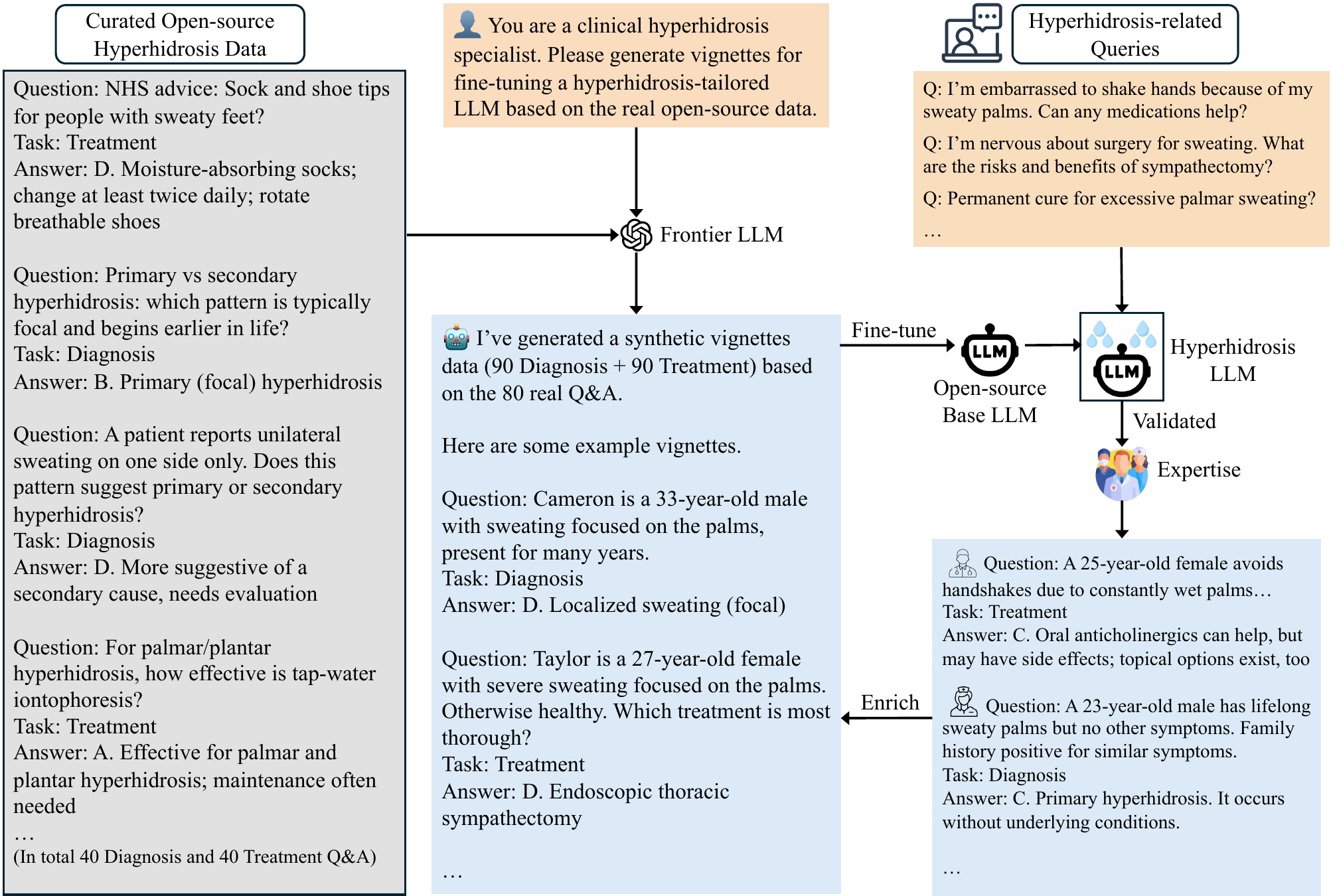}
  \caption {A demonstration of LLM4Sweat workflow.}
  \label{demo_llm4sweat}
\end{figure*}

We evaluated and demonstrated the performance of the proposed LLM4Sweat framework across diagnosis and treatment tasks under low-resource conditions. The workflow is present in Figure~\ref{demo_llm4sweat}. All models operated in a controlled evaluation environment, with prompt formats standardized across tasks to ensure fairness, and inference executed on a server with one NVIDIA A100 GPU to support efficient training, testing, and fair comparison.
\subsection{Datasets}

\subsubsection{Testing dataset}
We curated a balanced evaluation benchmark of 80 multiple-choice questions (40 diagnosis, 40 treatment) from open-access hyperhidrosis resources, including International Hyperhidrosis Society (IHHS)~\cite{ihhs}, National Health Service~\cite{NHS}, Mayo Clinic~\cite{mayo_dia_tre,mayo_overview}, DermNet~\cite{DermNet}, and MedlinePlus~\cite{MedlinePlus}. Questions cover both the diagnosis of hyperhidrosis and evidence-based treatment strategies, ensuring coverage of both clinical reasoning and therapeutic guidance. 

\subsubsection{Synthetic training dataset:}
To address data scarcity, we generated 180 synthetic vignettes (90 diagnosis, 90 treatment) using a frontier LLM (GPT-5) based on the real testing dataset. All generated cases were prompted and filtered for medical plausibility. This synthetic training dataset is for fine-tuning the open-source base LLMs before the inference and expert evaluation stage.

\subsubsection{Dynamic training dataset with validated clinical data:}

In addition to the static synthetic dataset, we incorporated iterative expert feedback cycles into the framework. In real-world applications, after inferences for real users' queries about hyperhidrosis, the model outputs are reviewed by domain experts, and validated responses are fed back into the dataset. In this experiment, we used the state-of-the-art LLM to represent specialists for practice. This process created a dynamic training corpus that improved alignment with clinical expertise and further fine-tuned the base model. For one cycle inference and expert evaluation, we got 40 outputs and corresponding 40 validated responses (20 for diagnosis and 20 for treatment) from specialists to enrich the synthetic training set, which can be used to fine-tune base models.

\subsection{Baselines and Our Models}

We evaluated our framework against open-source lightweight foundation models that had not been adapted to the hyperhidrosis domain: Llama-3.2-1B and Llama-3.2-3B. They served as baseline comparisons, reflecting the performance of general-purpose LLMs when directly applied to rare medical conditions without domain-specific fine-tuning.

In contrast, our proposed LLM4Sweat framework first fine-tuned these same base models on the dynamic synthetic training dataset composed of synthetic hyperhidrosis-related vignettes and also validated expert feedback. This enabled the models to learn domain-specific reasoning while maintaining computational efficiency.

\subsection{Performance Metrics}
For both diagnosis and treatment tasks, we evaluated models using four standard metrics: accuracy, precision, recall, and F1 scores for both diagnosis and treatment tasks. Accuracy provides an overall measure of correctness across all predictions, while Precision captures the proportion of model outputs that are clinically correct among all predicted positives. Recall quantifies the ability to identify true positives, a critical factor in medical contexts where missing a correct diagnosis or treatment option may have serious implications. The F1 score balances Precision and Recall, offering a more stable indicator of performance.

These metrics collectively allowed us to assess not only the correctness of model predictions but also their clinical robustness and reliability. In particular, Recall and F1 are especially important for hyperhidrosis-related tasks, where overlooking appropriate treatment guidance or misclassifying diagnosis categories could compromise patient care. By reporting all four measures, we provided a comprehensive evaluation of model performance and its potential for trustworthy clinical deployment.

\subsection{Hyperparameter Settings}

We trained the base models using LoRA-based supervised fine-tuning implemented in the Hugging Face PEFT library. 
This allowed us to efficiently adapt lightweight foundation models while keeping the number of trainable parameters small, 
making the framework more scalable for future deployment. We present other key hyperparameters here. The learning rate was tuned over $\{5 \times 10^{-6}, 5 \times 10^{-5}, 2 \times 10^{-4}, 1 \times 10^{-3}\}$, and training epochs were in the range of $\{1, 3, 5\}$ across models. A LoRA dropout rate of 0.05 was applied to improve generalization. For all the inference, we used the same temperature $=0.7$ and top-$p=0.9$ across all models.

These fine-tuning configurations and hyperparameter settings provided a balance between training stability, efficiency, and performance, 
enabling fair comparison between baselines and fine-tuned models under the LLM4Sweat framework.

\section{Results}
\label{Results}

\subsection{Comparison to Baselines}
Table~\ref{results} presents the performance of LLM4Sweat compared with raw baseline models on both diagnosis and treatment tasks. Baseline foundation models do not perform well under hyperhidrosis-specific evaluation, confirming that general-purpose LLMs struggle in rare, data-scarce medical domains. The smaller model, Llama-3.2-1B, achieves only 0.425 overall accuracy, with balanced but uniformly low performance in diagnosis (0.425) and treatment (0.425). The Llama-3.2-3B baseline shows modest improvements, particularly in treatment (0.700), but overall performance still remains insufficient (0.588). These results demonstrate that, without targeted fine-tuning, both LLMs fail to provide accurate medical decision support. Beyond accuracy, the baseline models also exhibit instability across precision, recall, and F1. For Llama-3.2-1B, precision remains moderate (0.543 overall), but both recall (0.421) and F1 (0.403) are very low, reflecting frequent failures to capture true positives and poor balance between sensitivity and specificity. Llama-3.2-3B achieves stronger precision (0.856 overall), particularly in treatment tasks (0.917), yet recall (0.587) and F1 (0.670) remain insufficient, indicating that while it can produce correct answers in some cases, it struggles to consistently identify them. These results also confirm that they fail to provide stable and clinically trustworthy performance across all evaluation metrics.

LLM4Sweat substantially outperforms baseline models across both tasks. After fine-tuning on the generated synthetic dataset, the 1B model achieves 0.925 diagnosis accuracy and 0.825 treatment accuracy, yielding an overall score of 0.875, more than doubling baseline accuracy. The 3B model improves from 0.588 to 0.900 overall accuracy, with diagnosis accuracy rising from 0.475 to 0.875 and treatment from 0.700 to 0.925. Importantly, gains extend beyond accuracy: both precision and recall increase, leading to F1 scores above 0.85 across all subtasks. This suggests that LLM4Sweat not only improves correctness but also achieves a better balance between sensitivity (capturing true positives) and specificity (avoiding false positives), both critical for clinical safety and trustworthiness.

Taken together, these findings validate the effectiveness of our data-centric fine-tuning strategy for our framework. By augmenting scarce real-world data with medically plausible synthetic cases and fine-tuning LLMs on the enriched data from expert evaluation, LLM4Sweat successfully transforms weak baseline models into high-performing, clinically relevant assistants. Last but not least, the results illustrate that even modestly sized LLMs, when fine-tuned carefully, can achieve competitive performance on specialized medical tasks, making them practical for deployment in resource-constrained environments.



  
  











  


\subsection{Ablation Studies}

To investigate the contribution of individual components, we first compare unadapted baseline models against their fine-tuned counterparts. As shown in Table~\ref{results}, for the 1B model, fine-tuning doubles performance, lifting overall accuracy to 0.813 and substantially improving both diagnosis (0.900 vs. 0.425) and treatment (0.725 vs. 0.425). Similarly, the 3B baseline achieves only 0.588 overall accuracy, with diagnosis accuracy at 0.475 and treatment at 0.700. After fine-tuning, the 3B model reaches 0.825 overall, 0.825 in diagnosis, and 0.825 in treatment, demonstrating that synthetic data augmentation and domain adaptation are the critical drivers of performance gains.

To further assess the contribution of expert-in-the-loop refinement, we compare fine-tuned models without and with expert evaluation. The results show that expert evaluation contributes measurable and consistent improvements across both model sizes. On the 1B model, expert refinement raises treatment accuracy from 0.725 to 0.825 and overall accuracy from 0.813 to 0.875, while maintaining high diagnosis accuracy (0.925). On the 3B model, diagnosis improves from 0.825 to 0.875, treatment from 0.825 to 0.925, and overall accuracy from 0.825 to 0.900. In both cases, F1 scores increase in parallel, indicating that expert validation enhances not just raw accuracy but also the stability and consistency of medical predictions.

These findings highlight two key insights of LLM4Sweat. First, fine-tuning with enriched synthetic data is an important driver of performance gains, elevating baseline models from near-random to clinically relevant levels. Second, expert-in-the-loop refinement provides the final layer of trustworthiness, ensuring that outputs are aligned with medical plausibility and clinical expectations. This mirrors real-world deployment needs: data-driven adaptation provides a strong foundation, but structured expertise remains essential for scaling applications.

\section{Discussions and Limitations}
\label{Discussions and Limitations}

The results highlight the effectiveness of the LLM4Sweat framework in addressing the challenges of hyperhidrosis support, a domain traditionally limited by scarce and noisy data. By combining generative augmentation with domain-specific fine-tuning and expert evaluation, the system achieved substantial performance improvements across multiple open-source base LLMs, with two Llama3.2 models fine-tuned reaching overall accuracy and strong gains in both diagnosis and treatment. Smaller variants, while starting from weaker baselines, benefited markedly from fine-tuning and further improved through inference-and-expert-evaluation cycles. These findings confirm that both data-centric (augmentation, fine-tuning) and process-centric (structured inference, expert validation) stages are essential to adapt general-purpose LLMs into trustworthy domain-specific medical assistants. 

Despite these promising outcomes, several limitations remain. First, the current evaluation relies on 80 curated multiple-choice questions, which, though carefully balanced, may not capture the full complexity of real-world patient interactions, including multi-modal symptoms, longitudinal histories, and emotional nuance. Second, expert evaluation was conducted in limited cycles; broader validation across diverse clinicians and patient populations is necessary to ensure robustness and fairness. Finally, deployment in clinical settings involves non-trivial challenges beyond technical accuracy: integration with electronic health records, regulatory approval, and alignment with ethical standards of care. Addressing these challenges will be essential to transition LLM4Sweat from a research-stage prototype into a clinically deployable decision support and patient guidance system, and the framework offers a blueprint for extending this approach to other rare and underrepresented medical conditions.

\section{Path to Deployment}
\label{Path to Deployment}

As demonstrated in Figure~\ref{demo_llm4sweat}, the LLM4Sweat workflow shows a clear pathway toward deployment in real-world applications. The following aspects are essential for a trustworthy and scalable deployment. 

\begin{itemize}
    \item \textbf{Expert Evaluation:}  
    The immediate step centers on rigorous expert-in-the-loop evaluation. In this work, it is done by the frontier LLM. In the future, the system will undergo structured evaluations with dermatologists, hyperhidrosis experts, and psychologists specializing in hyperhidrosis to assess diagnostic reliability, treatment appropriateness, and empathetic counseling quality. These evaluations will validate accuracy and guide iterative refinement of the fine-tuned model and data augmentation pipeline, ensuring that LLM4Sweat evolves into a clinically credible tool aligned with real-world reasoning and patient needs.  

    \item \textbf{Progressive Integration:}  
    Building on this foundation, LLM4Sweat can be gradually introduced into practice through both clinician-facing and patient-facing applications. For physicians, it may serve as a clinical decision support system, surfacing AI-generated suggestions alongside patient records while preserving clinician authority in decision-making. For patients, secure mobile or web interfaces could provide educational content, coping strategies, and personalized lifestyle guidance between consultations, and eventually extend into treatment recommendations under medical supervision. Deployment will follow established standard approval, medical constraints, and regulated certification.  

    \item \textbf{Generalization Ability:}  
    Beyond hyperhidrosis, the proposed methodology is transferable to other rare medical conditions, offering a scalable way to deploy trustworthy LLM-based assistants in underserved areas of medicine.  
\end{itemize}



\section{Conclusions}
\label{Conclusions}

We present LLM4Sweat, the first open-source LLM framework tailored for hyperhidrosis. By combining generative data augmentation, domain-specific fine-tuning, and expert-in-the-loop refinement, our approach addresses the challenges of data scarcity and trustworthiness. Experimental results across LLMs demonstrate that LLM4Sweat substantially outperforms raw baselines, achieving high accuracy in both diagnosis and treatment recommendation while improving stability and sensitivity, highlighting the necessity of both data-driven adaptation and expertise to ensure high and reliable performance.

Looking ahead, LLM4Sweat offers a generalizable and deployable method for applying LLMs in other rare medical conditions where curated data are insufficient but clinical need is high. Future work will also extend to richer patient cases, longitudinal scenarios, and multi-modal studies.

\bibliography{aaai2026}

\begin{thebibliography}{27}
\providecommand{\natexlab}[1]{#1}

\bibitem[{Alanezi(2024)}]{alanezi2024assessing}
Alanezi, F. 2024.
\newblock Assessing the effectiveness of ChatGPT in delivering mental health support: a qualitative study.
\newblock \emph{Journal of multidisciplinary healthcare}, 461--471.

\bibitem[{Barr et~al.(2025)Barr, Quan, Guo, and Sezgin}]{barr2025large}
Barr, A.~A.; Quan, J.; Guo, E.; and Sezgin, E. 2025.
\newblock Large language models generating synthetic clinical datasets: a feasibility and comparative analysis with real-world perioperative data.
\newblock \emph{Frontiers in Artificial Intelligence}, 8: 1533508.

\bibitem[{Bellapukonda, Mohan, and Sahu(2023)}]{bellapukonda2023logistic}
Bellapukonda, P.; Mohan, R. N. V.~J.; and Sahu, B. 2023.
\newblock Predicting Sweat Levels to Detect Hyperhidrosis: A Logistic Regression Approach.
\newblock In \emph{2023 14th International Conference on Computing, Communication and Networking Technologies (ICCCNT)}, 1--8. IEEE.

\bibitem[{Camargo et~al.(2024)Camargo, Gomes, Alves, and Corr{\^e}a}]{camargo2024polymer}
Camargo, F.~P.; Gomes, D.~R.; Alves, L.~R.; and Corr{\^e}a, C.~E. 2024.
\newblock Development of polymer insole in additive manufacturing for people with plantar hyperhidrosis.
\newblock \emph{Disciplinarum Scientia. S{\'e}rie: Naturais e Tecnol{\'o}gicas}, 25(2): 15--26.

\bibitem[{{DermNet}(2025)}]{DermNet}
{DermNet}. 2025.
\newblock Hyperhidrosis.
\newblock \url{https://dermnetnz.org/topics/hyperhidrosis/}.
\newblock Accessed: 2025-08.

\bibitem[{Farooq et~al.(2024)Farooq, Yao, Schukat, Little, and Corcoran}]{farooq2024derm}
Farooq, M.~A.; Yao, W.; Schukat, M.; Little, M.~A.; and Corcoran, P. 2024.
\newblock Derm-t2im: Harnessing synthetic skin lesion data via stable diffusion models for enhanced skin disease classification using vit and cnn.
\newblock In \emph{2024 46th Annual International Conference of the IEEE Engineering in Medicine and Biology Society (EMBC)}, 1--5. IEEE.

\bibitem[{Gregoriou et~al.(2019)Gregoriou, Sidiropoulou, Kontochristopoulos, and Rigopoulos}]{gregoriou2019management}
Gregoriou, S.; Sidiropoulou, P.; Kontochristopoulos, G.; and Rigopoulos, D. 2019.
\newblock Management strategies of palmar hyperhidrosis: challenges and solutions.
\newblock \emph{Clinical, cosmetic and investigational dermatology}, 733--744.

\bibitem[{Henning, Pedersen, and Jemec(2019)}]{henning2019genetic}
Henning, M.; Pedersen, O.; and Jemec, G. 2019.
\newblock Genetic disposition to primary hyperhidrosis: a review of literature.
\newblock \emph{Archives of dermatological research}, 311(10): 735--740.

\bibitem[{Henning et~al.(2021)Henning, Thorlacius, Ibler, and Jemec}]{henning2021diagnose}
Henning, M.~A.; Thorlacius, L.; Ibler, K.~S.; and Jemec, G.~B. 2021.
\newblock How to diagnose and measure primary hyperhidrosis: a systematic review of the literature.
\newblock \emph{Clinical Autonomic Research}, 31(4): 511--528.

\bibitem[{Hong, Lupin, and O'Shaughnessy(2012)}]{hong2012clinical}
Hong, C.-H.~H.; Lupin, M.; and O'Shaughnessy, K.~F. 2012.
\newblock Clinical evaluation of a microwave device for treating axillary hyperhidrosis.
\newblock \emph{Dermatologic Surgery}, 38(5): 728--735.

\bibitem[{Hyun et~al.(2023)Hyun, Kim, Im, Lee, and Kim}]{hyun2023machine}
Hyun, K.~Y.; Kim, J.~J.; Im, K.~S.; Lee, B.~S.; and Kim, Y.~J. 2023.
\newblock Machine learning analysis of primary hyperhidrosis for classification of hyperhidrosis type and prediction of compensatory hyperhidrosis.
\newblock \emph{Journal of Thoracic Disease}, 15(9): 4808--4817.

\bibitem[{{International Hyperhidrosis Society}(2025)}]{ihhs}
{International Hyperhidrosis Society}. 2025.
\newblock Hyperhidrosis.
\newblock \url{https://www.sweathelp.org/}.
\newblock Accessed: 2025-08.

\bibitem[{Kazeminia et~al.(2020)Kazeminia, Baur, Kuijper, van Ginneken, Navab, Albarqouni, and Mukhopadhyay}]{kazeminia2020gans}
Kazeminia, S.; Baur, C.; Kuijper, A.; van Ginneken, B.; Navab, N.; Albarqouni, S.; and Mukhopadhyay, A. 2020.
\newblock GANs for medical image analysis.
\newblock \emph{Artificial Intelligence in Medicine}, 109: 101938.

\bibitem[{Lin and Fang(2022)}]{lin2022systematic}
Lin, W.; and Fang, Y. 2022.
\newblock Primary Hyperhidrosis: A Systematic Review of Current Status and Potential Interventions.
\newblock In \emph{2022 6th International Conference on Universal Village (UV)}, 1--8. IEEE.

\bibitem[{Mart{\'\i}nez-Hern{\'a}ndez et~al.(2024)Mart{\'\i}nez-Hern{\'a}ndez, Estors-Guerrero, Galbis-Caravajal, Herv{\'a}s-Mar{\'\i}n, and Roig-Bataller}]{martinez2024endoscopic}
Mart{\'\i}nez-Hern{\'a}ndez, N.~J.; Estors-Guerrero, M.; Galbis-Caravajal, J.~M.; Herv{\'a}s-Mar{\'\i}n, D.; and Roig-Bataller, A. 2024.
\newblock Endoscopic thoracic sympathectomy for primary hyperhidrosis: an over a decade-long follow-up on efficacy, impact, and patient satisfaction.
\newblock \emph{Journal of Thoracic Disease}, 16(12): 8292.

\bibitem[{{Mayo Clinic}(2025{\natexlab{a}})}]{mayo_dia_tre}
{Mayo Clinic}. 2025{\natexlab{a}}.
\newblock Hyperhidrosis Diagnosis and Treatment.
\newblock \url{https://www.mayoclinic.org/diseases-conditions/hyperhidrosis/diagnosis-treatment/drc-20367173/}.
\newblock Accessed: 2025-08.

\bibitem[{{Mayo Clinic}(2025{\natexlab{b}})}]{mayo_overview}
{Mayo Clinic}. 2025{\natexlab{b}}.
\newblock Hyperhidrosis Overview.
\newblock \url{https://www.mayoclinic.org/diseases-conditions/hyperhidrosis/symptoms-causes/syc-20367152/}.
\newblock Accessed: 2025-08.

\bibitem[{{MedlinePlus}(2025)}]{MedlinePlus}
{MedlinePlus}. 2025.
\newblock Hyperhidrosis.
\newblock \url{https://medlineplus.gov/ency/article/007259.htm#:~:text=Hyperhidrosis%20is%20a%20medical%20condition,when%20they%20are%20at%20rest./}.
\newblock Accessed: 2025-08.

\bibitem[{{National Health Service}(2025)}]{NHS}
{National Health Service}. 2025.
\newblock Hyperhidrosis.
\newblock \url{https://www.nhs.uk/conditions/excessive-sweating-hyperhidrosis/}.
\newblock Accessed: 2025-08.

\bibitem[{Parashar, Adlam, and Potts(2023)}]{parashar2023impact}
Parashar, K.; Adlam, T.; and Potts, G. 2023.
\newblock The impact of hyperhidrosis on quality of life: a review of the literature.
\newblock \emph{American Journal of Clinical Dermatology}, 24(2): 187--198.

\bibitem[{Peng et~al.(2023)Peng, Yang, Chen, Smith, PourNejatian, Costa, Martin, Flores, Zhang, Magoc et~al.}]{peng2023study}
Peng, C.; Yang, X.; Chen, A.; Smith, K.~E.; PourNejatian, N.; Costa, A.~B.; Martin, C.; Flores, M.~G.; Zhang, Y.; Magoc, T.; et~al. 2023.
\newblock A study of generative large language model for medical research and healthcare.
\newblock \emph{NPJ digital medicine}, 6(1): 210.

\bibitem[{Qiu and Lan(2024)}]{qiu2024interactive}
Qiu, H.; and Lan, Z. 2024.
\newblock Interactive agents: Simulating counselor-client psychological counseling via role-playing llm-to-llm interactions.
\newblock \emph{arXiv preprint arXiv:2408.15787}.

\bibitem[{Sakhawat, Islam, and Reza(2024)}]{sakhawat2024patch}
Sakhawat, R.; Islam, M.~F.; and Reza, M.~T. 2024.
\newblock Design of a rectangular patch antenna-based electrolyte sensor for palmar hyperhidrosis patients.
\newblock In \emph{2024 27th International Conference on Computer and Information Technology (ICCIT)}, 3343--3347. IEEE.

\bibitem[{Singhal et~al.(2023)}]{singhal2023large}
Singhal, K.; et~al. 2023.
\newblock Large language models encode clinical knowledge.
\newblock \emph{Nature}, 620: 172--180.

\bibitem[{Solish, Benohanian, and Kowalski(2007)}]{solish2007impact}
Solish, N.; Benohanian, A.; and Kowalski, J.~W. 2007.
\newblock Impact of hyperhidrosis on quality of life and its assessment.
\newblock \emph{Dermatologic Clinics}, 25(4): 447--458.

\bibitem[{Strutton et~al.(2004)Strutton, Kowalski, Glaser, and Stang}]{strutton2004epidemiology}
Strutton, D.~R.; Kowalski, J.~W.; Glaser, D.~A.; and Stang, P.~E. 2004.
\newblock US prevalence of hyperhidrosis and impact on individuals with axillary hyperhidrosis: results from a national survey.
\newblock \emph{Journal of the American Academy of Dermatology}, 51(2): 241--248.

\bibitem[{Zhao et~al.(2025)Zhao, Bai, Bian, Chen, Li, Li, He, Yao, and Zhang}]{zhao2025autonomous}
Zhao, L.; Bai, J.; Bian, Z.; Chen, Q.; Li, Y.; Li, G.; He, M.; Yao, H.; and Zhang, Z. 2025.
\newblock Autonomous Multi-Modal LLM Agents for Treatment Planning in Focused Ultrasound Ablation Surgery.
\newblock \emph{arXiv preprint arXiv:2505.21418}.

\end{thebibliography}

\end{document}